# Character-focused Video Thumbnail Retrieval

Shervin Ardeshir [1]  Nagendra Kamath [1]  Hossein Taghavi [1]

## Abstract

We explore retrieving character-focused video frames as candidates for being video thumbnails. To evaluate each frame of the video based on the character(s) present in it, characters (faces) are evaluated in two aspects:

**Facial-expression**: We train a CNN model to measure whether a face has an acceptable facial expression for being in a video thumbnail. This model is trained to distinguish faces extracted from artworks/thumbnails, from faces extracted from random frames of videos.

**Prominence and interactions**: Character(s) in the thumbnail should be important character(s) in the video, to prevent the algorithm from suggesting non-representative frames as candidates. We use face clustering to identify the characters in the video, and form a graph in which the prominence (frequency of appearance) of the character(s), and their interactions (co-occurrence) are captured. We use this graph to infer the relevance of the characters present in each candidate frame.

Once every face is scored based on the two criteria above, we infer frame level scores by combining the scores for all the faces within a frame.

## 1. Introduction

Leveraging computer-vision and machine learning techniques, assistive tools could be developed to narrow down the search space for creators, when creating assets such as video thumbnails(Liu et al., 2015; Gao et al., 2009; Yuan et al., 2019). Here we evaluate the viability of such a task as a retrieval instance. Given a long format video, our goal is to rank the frames of the video in terms of their suitability of being a video thumbnail candidate. In the scope of this work we solely focus on character-focused candidates, i.e. we evaluate the suitability of each frame, using features extracted from the characters present in it.

[1]Netflix, Los Gatos, CA, USA. Correspondence to: Shervin Ardeshir <sardeshirbehrostaghi@netflix.com>, Nagendra Kamath <nkamath@netflix.com>, Hossein Taghavi <mtaghavi@netflix.com>.

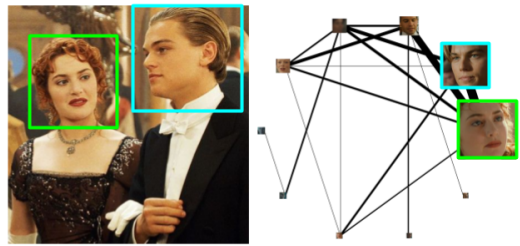

*Figure 1.* Left: candidate frame and the faces detected in it. Right: character graph of the source video containing the frame. Node sizes represent the prominence of the characters in the video, and the width of the edges encode the amount of interaction (co-occurrence) of two characters over the source video.

## 2. Framework

Given a frame $i$, we aim to assign a score $s_i$, capturing its suitability in terms of being a thumbnail image. We measure this score based on the characters(faces) present in frame $i$, namely $\{c_i^1, c_i^2, ..., c_i^k\}$. Our goal is to retrieve frames, in which the character(s) are primary characters, given the context of the video, and therefore representative of the video content. We measure this property for each character $j$ in frame $i$ of the video, and call it *prominence* score of character $c_i^j$, formalized as $p_i^j$. We also aim to suggest frames in which characters' facial expression is suitable for a thumbnail. Therefore for each character $c_i^j$, we estimate a facial expression score $e_i^j$.

To do the analysis mentioned above, we first need to decompose a frame into the character(s) present in it. We perform face detection(dli) on all the frames of the content. Then, each face is evaluated based on the two criteria mentioned above. Finally, we combine the scores for all the characters within a frame to infer a frame-level score, which we use for retrieval.

### 2.1. Facial Expression Model

We train a CNN for classifying whether a face has a thumbnail-suitable facial expression. To train such a model,

we extract faces from a dataset of video artworks/thumbnails (10k positive samples/faces) and a dataset of random video frames (10k negative samples).

**Architecture:** A convolutional neural network has been trained to classify whether a face has acceptable facial expression. The input to the network being the cropped face, the network has been trained with a binary cross-entropy loss with binary classification objective. We fine-tune the VGG-face (Parkhi et al., 2015) model, modifying the output layer to our binary classification task. The score predicted by this model on face $c_j^i$ is its facial expression score $e_j^i$

## 2.2. Character(s) prominence and interactions

Prominence is determined by performing face-detection on each frame(dli), extracting face-recognition embeddings(Parkhi et al., 2015), and clustering them to groups of faces belonging to different characters. Ideally, each cluster would include faces from a single character. Also, cluster size would capture the frequency of appearance of characters throughout the video, which we use as a proxy to the character's prominence. To do so, human faces in the frames of the video are detected (dli), face-recognition face embeddings (Parkhi et al., 2015) are then extracted from each of the faces, resulting in an embedding vector which is close to other faces of the same character.

Linkage-based clustering is applied to the embeddings, to group the faces belonging to the same identity. Each datapoint is initialized as a cluster, and every two clusters with members within a distance threshold of each other are merged. Given the nature of the features vectors extracted by the recognition network, this type of clustering could lead to a very high-purity clustering outcome, given a low enough linkage distance threshold (empirically set to .35 in our experiments). Threshold is selected based on the fact that the embeddings achieve near perfect precision for distances up to this threshold. The linkage based clustering effectively compensates for the low recall by making confident, iterative and transitive merging of datapoints.

**Inferring character prominence:** As each face in each frame is assigned to a cluster, and each cluster represents one of the characters in the video, the size of the cluster (frequency of appearance of character) is used as a proxy for character prominence. The prominence score of character $j$ in frame $i$, $p_i^j$ is inferred from the cluster to which it's face belongs. A cluster score is defined as its number of members (datapoints belonging to that cluster), normalized by the number of frames containing at least one face.

**Character interaction graph:** Once face clusters are identified, we form a video-level graph where each node is one of the face clusters (one of the characters). An example of the character graph has been shown in figure 1. The size of nodes encode the prominence of the characters as defined earlier. An edge between two nodes (visualizes as width of the edges) is used as a proxy for the amount of interactions two characters have. The weight $w_{kj}$ of an edge between node $k$ and $j$, is defined as the amount of co-occurrence that two nodes (characters) have in the source video, normalized by the number of frames containing at least one character.

## 2.3. Inferring frame-level scores

Once each face is evaluated in terms of expression suitability ($e_j^i$) and prominence ($p_j^i$), we assign an expression score and a relevance score to each frame of the video. Only taking into account prominence, the frame level relevance score for frame $i$ is calculated as a weighted mean of the face-level prominence scores, where weights are calculated based on the size of the face bounding box in the content of the frame. i.e. $p_i = \frac{\Sigma_j a_j^i p_j^i}{\Sigma_j a_j^i}$. Here $a_j^i$ is the area of face of character $j$ in frame $i$. Capturing both prominence and interaction for the characters in frame $i$, a more complete frame level relevance score could be calculated based on the sum of all the weights of the subgraph present in the image. Meaning $r_i = \Sigma_j(p_j^i + \Sigma_k^{j \neq k} w_{kj})$ or to be $p_i = \Sigma_j \frac{a_j^i}{\Sigma_j a_j^i}(p_j^i + \Sigma_k^{j \neq k} w_{kj})$. Similar to the frame-level prominence score, a frame-level expression score could also be calculated as: $e_i = \frac{\Sigma_j a_j^i e_j^i}{\Sigma_j a_j^i}$. The overall score is defined as a weighted sum of the expression score, and relevance (prominence and interaction) score. The linear weights are calculated empirically.

## 3. Experiments

We evaluate different components of the proposed approach in terms of standard retrieval metrics of mean average precision (mAP) and area under curve of the ROC curve (AUC-ROC) of the annotated thumbnail candidates.

**Dataset**: We collect a dataset of movies and linearly sample their frames (every 200 frame). We then ask annotators to label whether each frame is suitable as a thumbnail or not. We use 25% of the data, to tune weights for linearly combining the different relevance scores and facial expression scores using a grid search.

**Evaluation**: We evaluate the performance of different components of our approach in terms of standard retrieval metrics. We also evaluate baselines of randomly scoring the candidates, and simply assigning label 1 to all the frames containing characters.

# Character-focused Video Thumbnail Retrieval

|  | mAP | AUC-ROC |
| --- | --- | --- |
| random | 0.125 | 0.5 |
| only-faces | 0.1705 | 0.5208 |
| expression-model | 0.2505 | 0.5833 |
| prominence | 0.2332 | 0.6171 |
| interactions | 0.2143 | 0.5723 |
| prominence + interactions | 0.2242 | 0.5775 |
| combination | 0.2810 | 0.6327 |

Retrieval metrics on thumbnail retrieval task.

|  | mAP gain | AUC-ROC gain |
| --- | --- | --- |
| only-faces | 34.35% | 9.7% |
| expression-model | 97.39% | 24% |
| prominence | 83.76% | 31.26% |
| interactions | 68.87% | 21.74% |
| prominence + interactions | 76.67% | 22.84% |
| combination | 121.43% | 34.58% |

Gain in retrieval metrics.

**Future Work**: In the future, we aim to experiment with evolving each component to a more general purpose version of itself (not limited to human faces). We are also experimenting with an end-to-end training scheme, to alleviate the algorithm's dependency on heuristic assumptions.